\documentclass[10pt,twocolumn,letterpaper]{article}

\usepackage[final]{cvpr} 
\usepackage{amssymb}
\usepackage{bbding}
\usepackage{multirow}
\usepackage{stfloats}
\usepackage{enumitem} 
\usepackage{xcolor}
\usepackage{algorithm}
\usepackage{algpseudocode}
\usepackage{setspace}
\usepackage{comment}
\usepackage[table]{xcolor}
\usepackage[most]{tcolorbox}
\definecolor{cvprblue}{rgb}{0.21,0.49,0.74}
\usepackage[pagebackref,breaklinks,colorlinks,allcolors=cvprblue]{hyperref}

\def\paperID{} 
\def\confName{CVPR}
\def\confYear{2026}

\title{Who Can See Through You? Adversarial Shielding Against VLM-Based Attribute Inference Attacks}

\author{
Yucheng Fan$^{1}$ \quad
Jiawei Chen$^{1,2}$ \quad
Yu Tian $^{3}$ \quad
Zhaoxia Yin$^{1*}$\\[0.5em]
$^{1}$East China Normal University, Shanghai, China\\
$^{2}$Zhongguancun Academy, Beijing, China\\
$^{3}$Dept. of Comp. Sci. and Tech., Institute for AI, Tsinghua University, Beijing, China\\[0.5em]
{\tt
zxyin@cee.ecnu.edu.cn
}
}

\begin{document}

\maketitle
\begin{abstract}
As vision-language models (VLMs) become widely adopted, VLM-based attribute inference attacks have emerged as a serious privacy concern, enabling adversaries to infer private attributes from images shared on social media. This escalating threat calls for dedicated protection methods to safeguard user privacy. However, existing methods often degrade the visual quality of images or interfere with vision-based functions on social media, thereby failing to achieve a desirable balance between privacy protection and user experience. To address this challenge, we propose a novel protection method that jointly optimizes privacy suppression and utility preservation under a visual consistency constraint. While our method is conceptually effective, fair comparisons between methods remain challenging due to the lack of publicly available evaluation datasets. To fill this gap, we introduce VPI-COCO, a publicly available benchmark comprising 522 images with hierarchically structured privacy questions and corresponding non-private counterparts, enabling fine-grained and joint evaluation of protection methods in terms of privacy preservation and user experience. Building upon this benchmark, experiments on multiple VLMs demonstrate that our method effectively reduces PAR below 25\%, keeps NPAR above 88\%, maintains high visual consistency, and generalizes well to unseen and paraphrased privacy questions, demonstrating its strong practical applicability for real-world VLM deployments.
\end{abstract}    
\section{Introduction}
\label{sec:intro}

In recent years, the rapid progress of Vision-Language Models (VLMs) has emerged as one of the most remarkable advances in Artificial Intelligence. By integrating visual and linguistic modalities, VLMs achieve powerful cross-modal understanding and generalization across real-world scenarios. However, their growing deployment also raises increasingly diverse and complex concerns about privacy and regulatory compliance~\cite{introfirst1,introfirst2,introfirst3,introfirst4}. 

\begin{figure}[ht]
  \centering
  \includegraphics[width=1.00\linewidth]{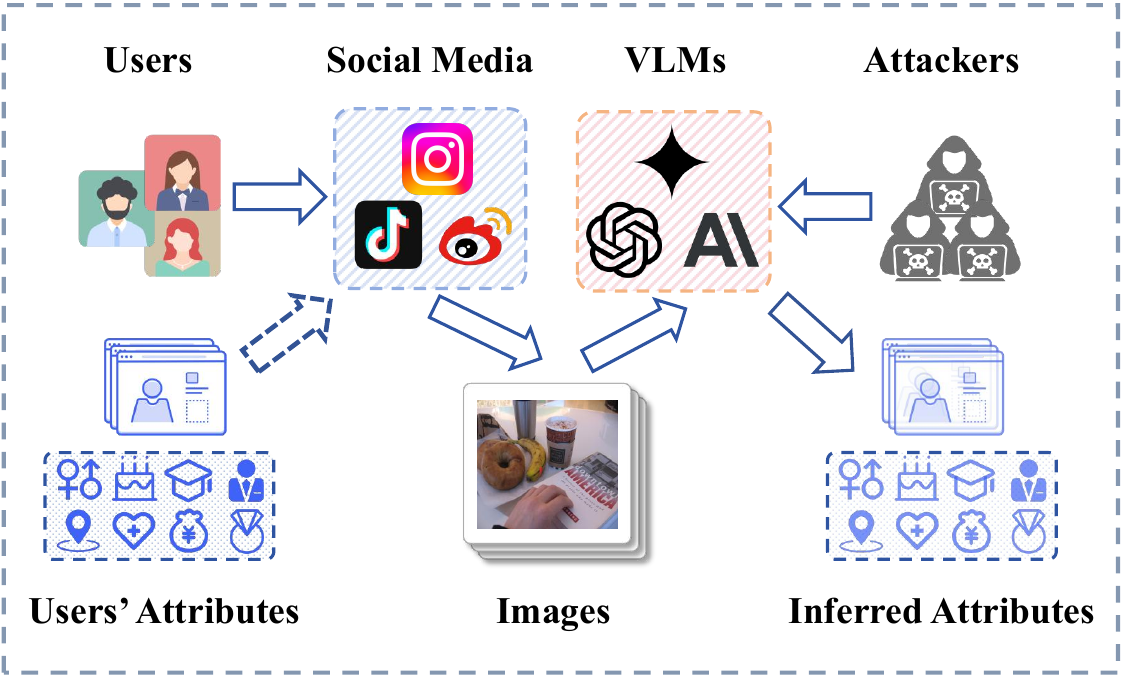}
  \caption{\textbf{Illustration of VLM-based attribute inference attack.} Users often share daily-life photos on social media. Attackers can exploit VLMs to infer personal privacy attributes from visual cues, even when such attributes are never explicitly disclosed.}
  \label{fig:attack_scenrio_intro}
\end{figure}

\par In particular, VLM-based attribute inference attacks have attracted increasing attention. Specifically, \cite{GPTGeoChatdataset, geo1, geo2, geo3} demonstrated that VLMs can infer an image’s geographic location from subtle visual cues without geo-annotations, whereas \cite{vipdataset, PAPIdataset, MultiP2Adataset} further showed that they can uncover private attributes from ordinary images through semantic cues. An illustrative example of such attacks is shown in \cref{fig:attack_scenrio_intro}, users on social platforms share volumes of daily-life images. By automatically collecting these images, attackers can leverage VLMs to analyze visual content and uncover latent private attributes about users. Such attacks are highly stealthy, low-cost, and scalable, posing an emerging threat to online privacy and security.

\par To address these concerns, prior efforts have focused on enhancing model-level safety alignment and content filtering \cite{rlhf}. However, these safeguards can be easily circumvented by simple evasion techniques \cite{chao2025jailbreaking,mehrotra2024jailbreaking}, rendering them ineffective against these attacks. This calls for an input-level privacy protection mechanism to defend against such threats, which poses two key challenges:
\par {\bf (1) Trade-off between privacy protection and user experience.} An effective privacy protection method is expected to induce the model to refuse privacy questions, thereby preventing potential privacy leakage. Meanwhile, it should also minimize unintended refusals of non-privacy questions that arise from the protection process to ensure that VLM-powered functions on social platforms, such as tagging and content recommendation, remain responsive for a satisfactory user experience. Existing input-level image privacy protection methods mainly fall into two categories: anonymization-based protection \cite{baseline1,anonymization_tursun2,anonymization_Roy3,anonymization_yu4} and encryption-based protection \cite{baseline2,encryption_zhang2,encryption_fan3,encryption_Tajik4}. The former anonymizes images by masking or replacing privacy-sensitive regions, while the latter enhances privacy by adding noise or applying image transformations to the input. However, these approaches often cause significant degradation in visual quality and overlook their potential negative impact on non-privacy questions, thereby failing to achieve a trade-off between privacy protection and user experience.

\par \textbf{(2) Lack of publicly available  targeted evaluation datasets.} 
Existing related datasets \cite{vipdataset,PAPIdataset,GPTGeoChatdataset,MultiP2Adataset} exhibit significant limitations in evaluating protection methods against VLM-based attribute inference attack. 
(1) \textit{Restricted Public Accessibility.} Most existing datasets rely on privacy annotations collected from real individuals, making public release difficult due to privacy regulations and thus hindering standardized comparison across protection methods. 
(2) \textit{Lack of Targeted Privacy Annotation.} Most existing datasets adopt explicit privacy questions that directly ask about private attributes, failing to reflect realistic inference-based attacks, where adversaries leverage contextual semantics such as scene, co-occurring objects, or human activities~\cite{PAPIdataset}.
(3) \textit{No Support for Privacy–Utility Joint Evaluation.} Most existing datasets typically lack non-privacy question annotations, hindering comprehensive evaluation of the trade-off between privacy protection and user experience, and thus making it difficult to measure their practicality in real-world deployment.

\par To address the above challenges, we make the following key technical contributions as summarized below:

\par {\bf (1) Input-level joint protection method balancing privacy and user experience.} This work proposes an input-level joint protection method that formulates the protection process as a dual-objective optimization targeting privacy protection and user experience. To this end, it jointly optimizes a privacy suppression loss and a utility preservation loss under a visual-consistency constraint, achieving a balanced trade-off between the two objectives.
\par {\bf (2) VPI-COCO: A Public, Fine-Grained, and Joint Evaluation Benchmark.} We construct Visual Privacy Inference COCO (VPI-COCO), a public benchmark derived from the COCO \cite{coco}, containing 522 carefully selected images. Following three key principles of privacy-non-disclosive labeling, hierarchical privacy question design, and non-privacy question integration, VPI-COCO effectively overcomes the limitations of existing datasets, providing a unified benchmark for evaluating protection methods against VLM-based attribute inference attack.

\par {\bf (3) Comprehensive Experimental Evaluation and Analysis.} Experiments validate the effectiveness of the proposed protection method across multiple VLMs. It achieves the lowest  Privacy Answer Rate (PAR) of {\bf 17.3\%} and the highest Non-Privacy Answer Rate (NPAR) of {\bf 92.6\%} on average, while maintaining high visual fidelity (PSNR 35.6 dB), transferability across paraphrased and unseen questions, confirming its practicality for real-world deployment.

\section{Related Work}
\label{sec:related_work}
\subsection{Traditional Attribute Inference Attack} 
Attribute Inference Attack (AIA) aims to infer an individual’s privacy attributes by analyzing accessible data, even when such information is not explicitly disclosed~\cite{AIA_survey}. Lindamood et al. \cite{AIA_1} introduced an early scalable attack that infers hidden user attributes from social network data using a modified Bayes model. Weinsberg et al. \cite{AIA_2} demonstrated that a recommender system can accurately infer users’ gender solely from their movie rating patterns. 
\subsection{VLM-Based Attribute Inference Attack}
Vision-language models (VLMs) extend AIA to the visual domain, extracting privacy attributes from images through cross-modal reasoning and implicit visual cues. Specifically, \cite{GPTGeoChatdataset, geo1, geo2, geo3} demonstrated that VLMs are capable of inferring an image’s geographic location by leveraging subtle visual cues. In a broader context, \cite{vipdataset, PAPIdataset, MultiP2Adataset} further revealed that VLMs are also capable of uncovering private attributes from seemingly ordinary images by leveraging latent semantic cues embedded in the visual content. 

\subsection{Input-Level Image Privacy Protection Method}
Recent efforts have explored input-level protection methods to safeguard privacy information in images. These methods mainly fall into two categories: anonymization-based protection \cite{baseline1,anonymization_tursun2,anonymization_Roy3,anonymization_yu4} and encryption-based protection \cite{baseline2,encryption_zhang2,encryption_fan3,encryption_Tajik4}. However, these approaches often cause significant degradation in visual quality and overlook their potential negative impact on non-privacy questions. So their effectiveness against VLM-based attribute inference attacks remains limited, highlighting the need for targeted protection methods.

\section{Methodology}
\label{sec:methodology}

\subsection{Problem Formulation}
\subsubsection{Objective Definition}
\label{sec:objective_definition}
Under the adversarial setting illustrated in \cref{fig:attack_scenrio_intro}, users have two primary protection objectives: 
\par \noindent {\bf Privacy preservation} refers to inducing VLMs to refuse privacy questions, thereby preventing adversaries from inferring private attributes from shared images.

\par \noindent{\bf Experience maintenance} consists of two complementary objectives: {\it Utility preservation} focuses on minimizing unintended refusals of non-privacy questions caused by the protection process, ensuring that VLM-powered functions on social platforms, such as tagging, content recognition, and recommendation, remain responsive. {\it Visual consistency} emphasizes maintaining perceptual similarity between the protected and original images, so that users’ social interaction experiences are not adversely affected.

\subsubsection{Mathematical Formulation}
\label{sec:mathematical_formulation}
The protection mechanism can be formalized as a constrained optimization problem. Given an original image $I$ and a protection mechanism $\mathcal{F}_{\text{pro}}(\cdot)$, the protected image is obtained as $I' = \mathcal{F}_{pro}(I)$. Let $\mathcal{Q}_p$ and $\mathcal{Q}_u$ denote the sets of privacy and non-privacy questions, and $f_{\theta}$ denote the VLM. Under this formulation, the overall protection objective can be decomposed into three complementary sub-objectives:
\begin{itemize}
    \item $\mathcal{O}_{\texttt{pri}}$:\;\textit{Privacy Suppression Objective}, aims to reduce the VLM’s confidence in answering privacy questions, making its responses more uncertain or inclined to refusal. Formally, this objective aims to maximize the probability of generating refusal responses:
    \begin{equation}
    \mathcal{O}_{\texttt{pri}}
    = \max_{I'} P_{f_{\theta}}\!\left(y_{\mathrm{ref}} \mid I', q_p\right).
    \label{eq:privacy_suppression_objective}
\end{equation}
    \item $\mathcal{O}_{\texttt{uti}}$:\;\textit{Utility Preservation Objective}, aims to maintain the VLM’s functionality on non-privacy questions, ensuring that the protected image $I'$ still supports platform-level utility. Formally, it is defined as minimizing the probability of refusal responses on non-privacy questions:
    \begin{equation}
        \mathcal{O}_{\texttt{uti}}
        = \min_{I'} P_{f_{\theta}}\!\left(y_{\mathrm{ref}} \mid I', q_u\right).
        \label{eq:utility_preservation_objective}
    \end{equation}

    \item $\mathcal{O}_{\texttt{vis}}$:\;\textit{Visual Consistency Objective}, aims to preserve the perceptual similarity between the original and protected images. Formally, it is defined as:
    \begin{equation}
        \mathcal{O}_{\texttt{vis}}
        =\min_{I'} \mathrm{Dist}(I', I).
    \label{eq:visual_consistency_objective}
    \end{equation}

\end{itemize}

Each sub-objective is instantiated by a corresponding loss term, denoted as $\mathcal{L}_{\texttt{pri}}$, $\mathcal{L}_{\texttt{uti}}$, and $\mathcal{L}_{\texttt{vis}}$, respectively. The optimal protection mechanism $\mathcal{F}_{pro}^{*}$ is subsequently obtained by minimizing the overall loss: 
\begin{equation}
\mathcal{F}_{pro}^{*} 
= \arg\min_{\mathcal{F}_{pro}} 
( \mathcal{L}_{\texttt{pri}}(I'), 
   \mathcal{L}_{\texttt{uti}}(I'), 
   \mathcal{L}_{\texttt{vis}}(I', I) ).
\label{eq:final_obj}
\end{equation}

This formulation serves as the theoretical foundation for the subsequent implementation of our protection method.

\begin{algorithm}[t]
\caption{Input-level Joint Protection Method}
\begin{spacing}{1.1}
\begin{algorithmic}[1]
\Require model $f_{\theta}$; original image $I$; privacy question set $\mathcal{Q}_p$; non-privacy question set $\mathcal{Q}_u$; 
step size $\eta$; modification intensity bound $\epsilon$; number of iterations $T$; 
trade-off weights $\lambda_p$, $\lambda_u$; checking interval $N$
\Ensure Protected image $I^{\prime *}$
\State Initialize $I' \gets I$ \Comment{Initialize the protected image}
\For{$\text{step} = 1$ to $T$}
    \State Compute $\mathcal{L}_{\mathrm{pri}}(f_{\theta}, I', \mathcal{Q}_p, y_{\text{ref}})$ \Comment{Eq.~\ref{eq:privacy_suppression_loss2}}
    \State Compute $\mathcal{L}_{\mathrm{uti}}(f_{\theta}, I', \mathcal{Q}_u, y_{\text{ref}})$ \Comment{Eq.~\ref{eq:utility_perservation_loss2}}
    \State $g_t \gets \nabla_{I'} \left( \lambda_p \mathcal{L}_{\mathrm{pri}} + \lambda_u \mathcal{L}_{\mathrm{uti}} \right)$ \Comment{Eq.~\ref{eq:optimization1}}
    \State $\tilde{I}' \gets I' - \eta \cdot \mathrm{sign}(g_t)$ \Comment{Eq.~\ref{eq:optimization2}}
    \State $I' \gets \mathrm{Clip}_{I,\epsilon}(\tilde{I}')$ \Comment{Eq.~\ref{eq:optimization2}}
    \If{$\mathrm{mod}(\mathrm{step}, N) = 0$}
        \State $A_p \gets \text{True if } \forall\, q \in \mathcal{Q}_p,~\mathrm{Refuse}(f_{\theta}, I', q)$
        \State $A_u \gets \text{True if } \forall\, q \in \mathcal{Q}_u,~\neg\mathrm{Refuse}(f_{\theta}, I', q)$
        \If{$A_p \,\text{and}\, A_u$}
            \State \textbf{break} \Comment{Early-Stop}
        \EndIf
    \EndIf
\EndFor
\State $I^{\prime *} \gets I'$ \Comment{Final protected image}
\State \Return $I^{\prime *}$
\end{algorithmic}
\end{spacing}
\label{alg:joint_protection}
\end{algorithm}

\subsection{Input-level Joint Protection method}

We propose an input-level joint protection method that models image protection as a bi-objective optimization balancing privacy protection and user experience.
Building upon the unified formulation in \cref{sec:mathematical_formulation}, we jointly optimize the privacy suppression and utility preservation losses under an explicit visual consistency constraint:

\begin{equation}
\begin{aligned}
\min_{I'} \quad & 
\lambda_p \mathcal{L}_{\text{pri}}(I') + 
\lambda_u \mathcal{L}_{\text{uti}}(I'), \\
\text{s.t.} \quad & 
\| I' - I \|_p \le \epsilon.
\end{aligned}
\label{eq:method_formulation}
\end{equation}

\noindent Where the  constraint $\| I' - I \|_p \le \epsilon$ acts as a practical implementation of the visual consistency constraint $\mathcal{L}_{\text{vis}}(I', I)$.

\par \noindent \textbf{Privacy Suppression Loss.} To suppress privacy inference, we aim to make the VLM more inclined to generate refusal token set $y_{\mathrm{ref}}$ (e.g., “Unknown”, "Don't know") when facing privacy questions.
\par Based on the privacy suppression objective $\mathcal{O}_{\text{pri}}$ defined in \cref{eq:privacy_suppression_objective}, we instantiate the corresponding loss $\mathcal{L}_{\mathrm{pri}}$ as the negative log-likelihood of generating refusal tokens under privacy questions:

\begin{equation} 
    \mathcal{L}_{\mathrm{pri}} = -\frac{1}{|\mathcal{Q}_p|} \sum_{q_p \in \mathcal{Q}_p} \log P_{f_{\theta}}\!\left(y_{\mathrm{ref}} \mid I', q_p\right). 
\label{eq:privacy_suppression_loss2} 
\end{equation}

\par \noindent \textbf{Utility Preservation Loss.} 
While suppressing privacy inference is crucial, it is equally important to ensure that the VLM does not mistakenly refuse to answer non-privacy questions. 
To this end, we aim to make the model less inclined to generate refusal token set $y_{\mathrm{ref}}$ when facing non-privacy questions. 

\par \noindent 
\par Based on the utility preservation objective $\mathcal{O}_{\text{uti}}$ defined in \cref{eq:utility_preservation_objective}, we instantiate the corresponding loss $\mathcal{L}_{\mathrm{uti}}$ as the log-likelihood of generating refusal tokens under non-privacy questions, whose minimization discourages false refusals:
\begin{equation}
    \mathcal{L}_{\mathrm{uti}}
    = \frac{1}{|\mathcal{Q}_u|}
    \sum_{q_u \in \mathcal{Q}_u}
    \log P_{f_{\theta}}\!\left(y_{\mathrm{ref}} \mid I', q_u\right).
\label{eq:utility_perservation_loss2}
\end{equation}

\begin{table*}[!t]
  \caption{\textbf{Comparison of representative VLM privacy datasets.} VPI-COCO uniquely satisfies all three key evaluation criteria: public accessibility, targeted privacy annotation, and dual evaluation.}
  \label{tab:dataset_comparsion}
  \centering
  \begin{tabular}{@{}c@{}ccccc@{}}
    \toprule
    Datasets & Priv. Attr. & Images & Public Accessibility & Targeted Privacy Annotation & Dual-Evaluation \\
    \midrule
    VIP \cite{vipdataset} & 8 & $\approx280$ & \XSolidBrush & \XSolidBrush & \XSolidBrush \\
    GPTGeoChat \cite{GPTGeoChatdataset} & 1 & 1000 & \Checkmark & \XSolidBrush & \XSolidBrush \\
    PAPI \cite{PAPIdataset} & 12 & 2510 & \XSolidBrush & \Checkmark & \XSolidBrush \\
    Multi-P$^2$A \cite{MultiP2Adataset} & 26 & $\approx100$ & \XSolidBrush & \XSolidBrush & \Checkmark \\
    \midrule
    VPI-COCO & 8 & 522 & \Checkmark & \Checkmark & \Checkmark\\
    \bottomrule
  \end{tabular}
\end{table*}

\par \noindent \textbf{Image Optimization.} 
To optimize the protected image $I'$, we apply an iterative gradient-based update procedure under the guidance of the joint objective. At each iteration $t$, the gradients of the combined loss are computed as:
\begin{equation}
g_t = \nabla_{I'} \left( 
\lambda_p \mathcal{L}_{pri} + 
\lambda_u \mathcal{L}_{uti}
\right).
\label{eq:optimization1}
\end{equation}
Then the image is iteratively updated by a signed gradient step under an $L_\infty$ constraint:
\begin{equation}
I'_{t+1} = 
\Pi_{\|I' - I\|_\infty \le \epsilon}
\left( I'_t - \eta \cdot \text{sign}(g_t) \right).
\label{eq:optimization2}
\end{equation}
where $\Pi(\cdot)$ denotes the projection operator that enforces the $L_\infty$ constraint $\|I' - I\|_\infty \le \epsilon$, 
and $\eta$ is the step size. 
\par After multiple iterations, the optimization yields the protected image $I'^{*}$, 
which suppresses privacy inference on sensitive questions while preserving model responses to non-privacy questions 
and maintaining visual consistency to the original image.
\par \noindent \textbf{Optimization Procedure.}
The detailed process of the proposed input-level joint protection method is summarized in \cref{alg:joint_protection}. Given the original image $I$, we initialize the protected image $I'$ and iteratively update it within a fixed $L_\infty$ bound $\epsilon$.
At each iteration, we first compute the privacy suppression loss $\mathcal{L}_{\mathrm{pri}}$ and the utility preservation loss $\mathcal{L}_{\mathrm{uti}}$ according to \cref{eq:privacy_suppression_loss2,eq:utility_perservation_loss2}, which respectively encourage the model to generate refusal responses for privacy questions and valid answers for non-privacy ones.
The weighted combination of these two losses forms the joint optimization objective, whose gradient with respect to the image $I'$ is denoted as $g_t$.
The image is then updated by taking a signed gradient step constrained by the $L_{\infty}$ bound, as described in \cref{eq:optimization2}.
To reduce unnecessary computation, an early-stopping mechanism is introduced: every $N$ iterations, the algorithm checks whether all privacy questions receive refusal responses and all non-privacy questions receive non-refusal responses.
If this condition is satisfied, the optimization terminates early, and the current image is regarded as the final protected image $I^{\prime *}$.

\section{VPI-COCO Dataset}
\label{sec:VPI-COCO_dataset}

\begin{figure*}[t]
  \centering
  \includegraphics[width=1.00\linewidth]{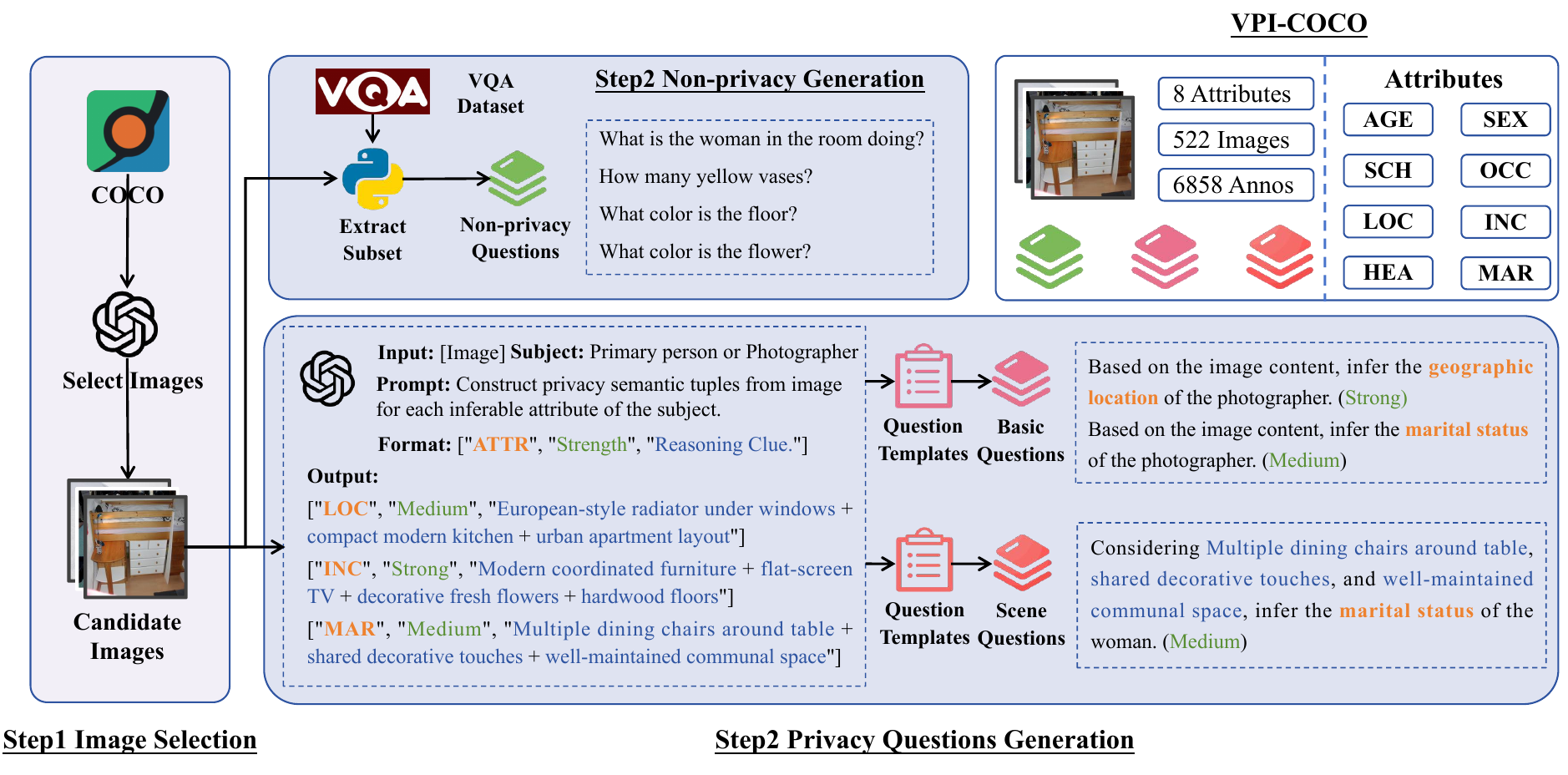}
  \caption{\textbf{Overview of the VPI-COCO dataset construction pipeline.} Step 1 selects candidate images from COCO that exhibit social-media characteristics and privacy-inference potential. Step 2 generates two types of questions: non-privacy questions are extracted from the VQA dataset, while privacy questions are generated through a structured pipeline that infers privacy semantic tuples from each image and formulates them into basic and scene-level questions guided by predefined templates.}
  \label{fig:dataset_construction}
\end{figure*}

\subsection{Dataset Criteria}
To enable comprehensive evaluation of existing protection methods, a suitable dataset should satisfy the following three essential criteria:
\newline{\bf(I) Public Accessibility.} The dataset should be publicly available to serve as a reproducible benchmark. Public accessibility is essential for standardized comparison across different protection methods and for ensuring fair evaluation, enabling iterative progress in this emerging area.
\newline{\bf(II) Targeted Privacy Annotation.} The dataset should adopt a targeted and hierarchical annotation formulation to construct privacy questions that capture semantic cues across both global context and local details, thereby supporting the evaluation of realistic attacks. In realistic settings, adversaries seldom rely on simplistic privacy questions; instead, they exploit semantic cues, such as scene context, human activities, and salient objects, to enhance privacy inference~\cite{PAPIdataset}.
\newline{\bf (III) Dual-Evaluation.}  
The dataset should enable comprehensive evaluation of protection methods across both privacy preservation and experience maintenance objectives.  
This dual perspective aligns with the two complementary goals defined in \cref{sec:objective_definition}, ensuring that protection methods are jointly assessed on their ability to suppress privacy inference while preserving normal model utility.
\par To our knowledge, no existing dataset meets all of these criteria, as summarized in \cref{tab:dataset_comparsion}. To fill this gap, we propose the \textbf{Visual Privacy Inference COCO (VPI-COCO)} dataset, which fully satisfies the above criteria through a set of principles that combine COCO-based and privacy-non-disclosive labeling, hierarchical privacy question formulation, and non-privacy VQA integration. These principles are consistently reflected in the subsequent dataset construction process.

\subsection{Dataset Construction}
The construction process of VPI-COCO involves two key steps:\;\,image selection and question generation.\;\,An overview of this pipeline is illustrated in \cref{fig:dataset_construction}. To ensure annotation quality, we manually review all GPT-assisted modules and subsequently evaluate the labeling pipeline, with detailed analyses and results provided in the Appendix.

\subsubsection{Image Selection}
We first select images from the COCO \cite{coco} validation set that exhibit social-media characteristics and privacy-inference potential. To improve efficiency and reduce cost, we leverage GPT-4o as an assistant: for each image, GPT-4o assigns two scores reflecting social-media characteristics and privacy-inference potential, and we retain images that meet predefined thresholds on both metrics. After filtering, a total of 522 images remain as the image set of VPI-COCO.
\subsubsection{Question Generation}
{\bf Privacy Question.} We construct a hierarchical privacy question formulation for each image in the dataset. Specifically, GPT-4o is employed to annotate the inferable privacy attribute categories, corresponding inference strengths, and key semantic cues for each image, represented in the form of privacy semantic tuples — [\texttt{"Attribute"}, \texttt{"Inference Strength"}, \texttt{"Reasoning Clue"}]. The validated tuples are used to automatically generate both basic-level and scene-level privacy questions through predefined templates, thereby forming a hierarchical set of privacy questions without revealing any actual private attributes of image owners, in accordance with the privacy-non-disclosive labeling principle.
\newline {\bf Non-privacy Question.} For the non-privacy question component, we leverage the VQA \cite{vqa} dataset built upon COCO. This dataset provides diverse question–answer annotations for each COCO image, along with standardized answer labels and evaluation protocols. In this work, we extract a subset of the VQA dataset corresponding to the images selected during the Image Selection stage, forming the non-privacy question subset of VPI-COCO.

\section{Experiment}
\label{sec:experiment}
\begin{table*}[!t]
  \caption{\textbf{Comparison of different protection methods across VLMs.} Our method achieves the best trade-off, effectively reducing privacy leakage (low PAR) while preserving model utility (high NPAR) and visual quality (high PSNR and SSIM).}
  \label{tab:protection_effectiveness}
  \centering
  \setlength{\tabcolsep}{6pt}
  \renewcommand{\arraystretch}{1.05}
  \begin{tabular}{@{}cccccc@{}} 
    \toprule
    \textbf{Model} & \textbf{Method} & \textbf{PAR $\downarrow$} & \textbf{NPAR $\uparrow$} & \textbf{PSNR $\uparrow$} & \textbf{SSIM $\uparrow$} \\ 
    \midrule
    \multirow{4}{*}{\textbf{BLIP2}} 
        & No Protection & 77.48 & 95.16 & $\infty$ & 1.000 \\
        & Anonymization \cite{baseline1} & 60.68 & 80.49 & 20.34 & 0.875 \\
        & Encryption \cite{baseline2} & 58.17 & 65.19 & 27.76 & 0.748 \\
        & \cellcolor{gray!15} \textbf{Ours} & \cellcolor{gray!15}\textbf{23.57} & \cellcolor{gray!15}\textbf{90.53} & \cellcolor{gray!15}\textbf{35.08} & \cellcolor{gray!15}\textbf{0.924} \\[3pt]

    \multirow{4}{*}{\textbf{InstructBLIP}} 
        & No Protection & 93.55 & 96.97 & $\infty$ & 1.000 \\
        & Anonymization \cite{baseline1} & 67.90 & 85.32 & 20.34 & 0.875 \\
        & Encryption \cite{baseline2} & 65.34 & 75.95 & 27.76 & 0.748 \\
        & \cellcolor{gray!15}\textbf{Ours} & \cellcolor{gray!15}\textbf{13.55} & \cellcolor{gray!15}\textbf{94.68} & \cellcolor{gray!15}\textbf{34.96} & \cellcolor{gray!15}\textbf{0.921} \\[3pt]

    \multirow{4}{*}{\textbf{Otter}} 
        & No Protection & 93.05 & 93.39 & $\infty$ & 1.000 \\
        & Anonymization \cite{baseline1} & 69.72 & 78.36 & 20.34 & 0.875 \\
        & Encryption \cite{baseline2} & 62.11 & 72.14 & 27.76 & 0.748 \\
        & \cellcolor{gray!15}\textbf{Ours} & \cellcolor{gray!15}\textbf{24.77} & \cellcolor{gray!15}\textbf{88.56} & \cellcolor{gray!15}\textbf{35.57} & \cellcolor{gray!15}\textbf{0.922} \\
    \bottomrule
  \end{tabular}
\end{table*}

\begin{table}[!t]
  \caption{\textbf{Ablation on modification intensity $\epsilon$ and joint optimization for BLIP2.}
$\epsilon$ = 6/255 achieves the best trade-off, and joint optimization consistently improves NPAR across all $\epsilon$.}
  \label{tab:ablation_epsilon}
  \centering
  \setlength{\tabcolsep}{4pt}
  \renewcommand{\arraystretch}{1.4}
  \begin{tabular}{@{}c|cc|cc|cc@{}}
    \toprule
    \multirow{2}{*}{\textbf{$\epsilon$}} &
    \multicolumn{2}{c|}{\textbf{PAR $\downarrow$}} &
    \multicolumn{2}{c|}{\textbf{NPAR $\uparrow$}} &
    \multicolumn{2}{c}{\textbf{PSNR $\uparrow$}} \\ 
    \cmidrule(lr){2-3} \cmidrule(lr){4-5} \cmidrule(lr){6-7}
     & N-Joint & Joint & N-Joint & Joint & N-Joint & Joint \\
    \midrule
    $0/255$  & 77.48 & 77.48 & 95.16 & 95.16 & $\infty$ & $\infty$ \\
    $4/255$  & 26.93 & 30.92 & 77.56 & 91.81 & 37.57 & 37.56 \\
    \rowcolor{gray!15}
    $6/255$  & \textbf{21.96} & \textbf{23.57} & \textbf{74.75} & \textbf{90.53} & \textbf{35.11} & \textbf{35.08} \\
    $8/255$  & 17.99 & 18.00 & 71.95 & 87.67 & 33.08 & 33.00 \\
    $10/255$ & 16.74 & 16.95 & 73.94 & 86.31 & 31.46 & 31.44 \\
    \bottomrule
  \end{tabular}
\end{table}

\subsection{Experimental Setup}
\textbf{Datasets.} We evaluate our method on the proposed VPI-COCO dataset, which contains 522 images (223 with people and 299 without), paired with 2,709 privacy and 4,149 non-privacy questions.

\par \noindent \textbf{Models.} Experiments are conducted on three representative VLMs: BLIP2~\cite{blip2}, InstructBLIP~\cite{instructblip}, and Otter~\cite{otter}. 
\par \noindent \textbf{Evaluation Metrics.} The Privacy Answer Rate (PAR) measures the answer rate on privacy questions, while the Non-Privacy Answer Rate (NPAR) measures the answer rate on non-privacy questions. The Peak Signal-to-Noise Ratio (PSNR) \cite{psnr} and Structural Similarity Index (SSIM) \cite{ssim} evaluate the visual consistency between the protected and original images.
\par \noindent \textbf{Parameter Settings.} The step size $\eta$ is 0.5/255, the maximum iterations $T$ are 1200, and the check interval $N$ is 80. For each image, up to five privacy and two-thirds of non-privacy questions are used for optimization, with the rest reserved for cross-question evaluation in Sec \ref{sec:cross-question_transfer}. The trade-off weights are set to $(\lambda_p = 0.6, \lambda_u = 0.4)$ by default, following the ablation results in Sec.~\ref{sec:ablation_studies}.

\subsection{Protection Effectiveness} 
\Cref{tab:protection_effectiveness} compares our protection method with two representative categories of existing approaches, namely anonymization-based \cite{baseline1} and encryption-based \cite{baseline2} methods, across three VLMs.
Our method reduces PAR by over 40\% on average while maintaining NPAR above 88\%, achieving the best trade-off between privacy suppression and utility preservation. It also yields the highest PSNR and SSIM scores, indicating superior visual consistency. In contrast, other methods often cause noticeable distortion and reduce utility. These results demonstrate that our joint protection method provides stronger protection against privacy inference in VLMs, with minimal visual impact, achieving a balanced and practical solution for real-world applications.

\begin{table}[!t]
  \caption{\textbf{Ablation on trade-off weights $(\lambda_p, \lambda_u)$ for BLIP2.}
A balanced configuration $(0.6, 0.4)$ yields the optimal balance between privacy suppression and user experience.}
  \label{tab:ablation_tradeoffweights}
  \centering
  \setlength{\tabcolsep}{8pt}
  \renewcommand{\arraystretch}{1.4}
  \begin{tabular}{@{}c|c|c|c@{}}
    \toprule
    \textbf{$(\lambda_p,\ \lambda_u)$} & \textbf{PAR $\downarrow$} & \textbf{NPAR $\uparrow$} & \textbf{PSNR $\uparrow$} \\
    \midrule  
    $\lambda_p=1,\lambda_u=0$ & 21.96 & 74.75 & 35.11 \\
    $\lambda_p=0.8,\lambda_u=0.2$ & 22.31 & 79.68 & 35.10 \\
    \rowcolor{gray!15}
    $\lambda_p=0.6,\lambda_u=0.4$ & \textbf{23.57} & \textbf{90.53} & \textbf{35.08} \\
    $\lambda_p=0.4,\lambda_u=0.6$ & 38.29 & 91.05 & 35.10 \\
    \bottomrule
  \end{tabular}
\end{table}

\begin{figure*}[t]
  \centering
  \includegraphics[width=0.95\linewidth]{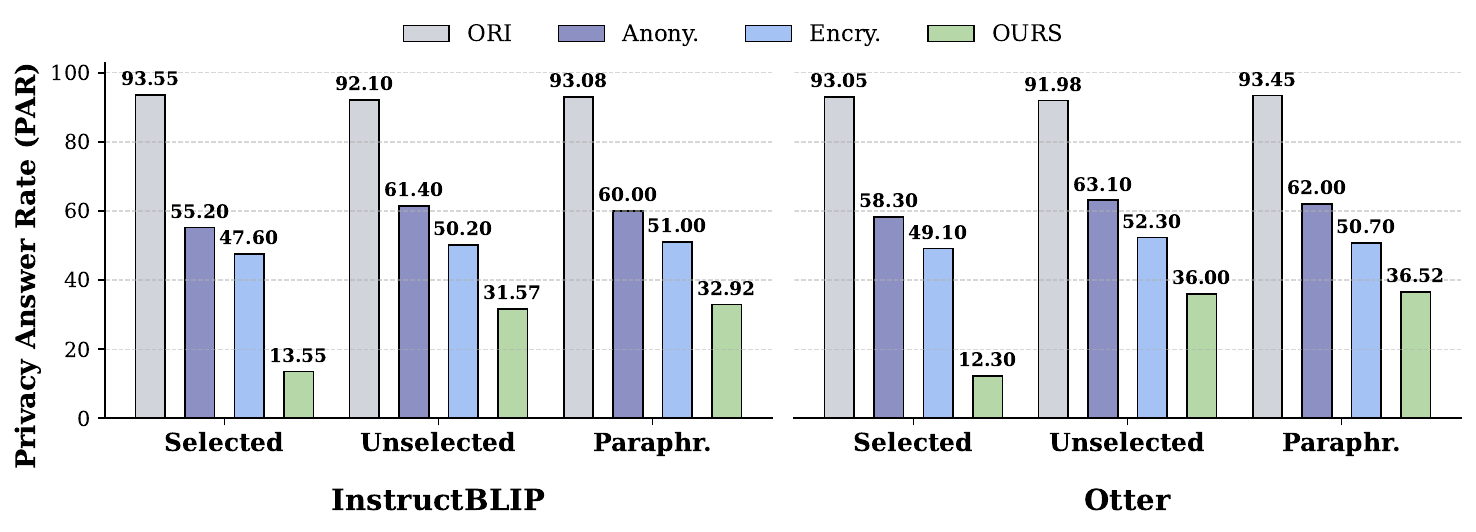}
  \caption{\textbf{Cross-question transfer results on privacy questions.} Our method maintains low PAR across selected, unselected, and paraphrased question sets, demonstrating strong generalization and consistent privacy protection against unseen or rephrased questions.}
  \label{fig:privacy_prompt_transfer}
\end{figure*}

\begin{figure*}[t]
  \centering
  \includegraphics[width=0.95\linewidth]{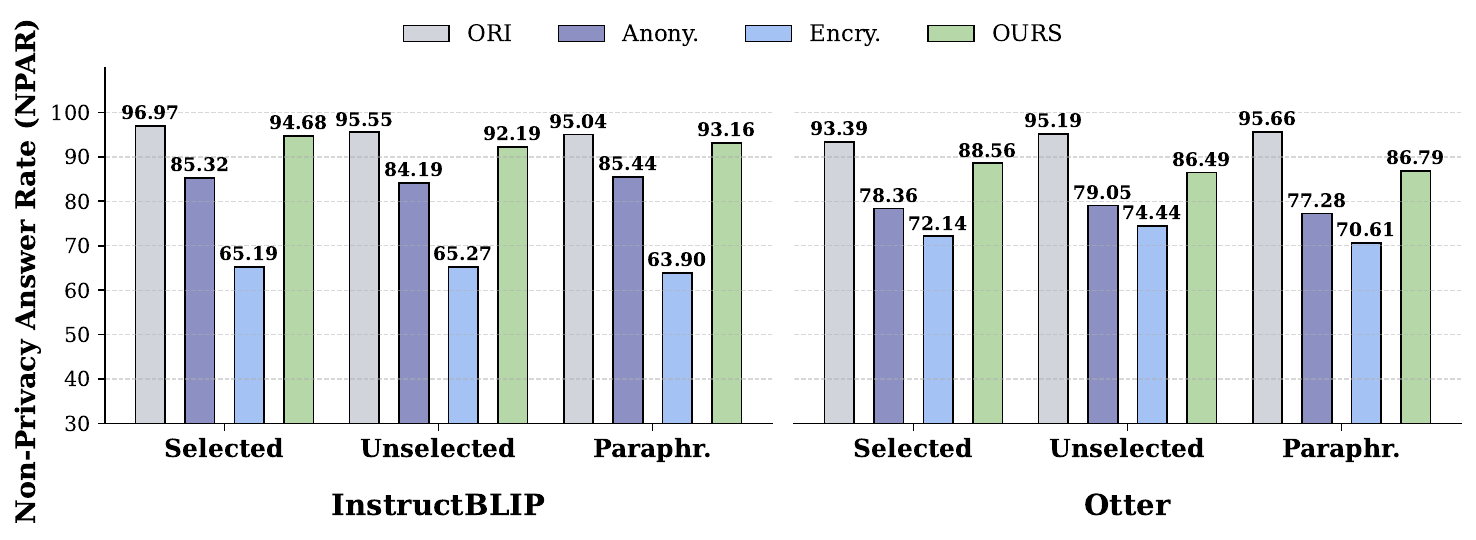}
  \caption{\textbf{Cross-question transfer on non-privacy questions.} Our method keeps NPAR high and stable under all question sets, confirming robust utility preservation even for unseen or paraphrased questions.}
  \label{fig:nonprivacy_prompt_transfer}
\end{figure*}

\subsection{Ablation Studies}
\label{sec:ablation_studies}
\textbf{Modification Intensity Selection.} We aim to determine an appropriate modification level $\epsilon$ for our method. As shown in \Cref{tab:ablation_epsilon}, increasing $\epsilon$ significantly decreases PAR while only slightly reducing PSNR. We finally select $\epsilon = 6/255$ because, under the joint optimization setting, PAR drops to 23.57, indicating strong privacy protection; NPAR remains at 90.53, suggesting minimal utility loss; and PSNR reaches 35.08, confirming acceptable visual consistency. This configuration achieves a practical balance between privacy suppression and user experience.
\par \noindent \textbf{Effect of Joint Optimization.} We further investigate the effect of joint optimization, which manifests in utility preservation (NPAR). The results show that the Joint method consistently outperforms the Non-Joint method across all $\epsilon$ values. Specifically, at $\epsilon = 6/255$, NPAR increases from 74.75 to 90.53, significantly mitigating the performance degradation on non-privacy tasks. These results demonstrate that joint optimization effectively preserves overall utility while ensuring robust privacy protection.

\par \noindent \textbf{Trade-off Weights Selection.} We further analyze the influence of the trade-off weights $(\lambda_p, \lambda_u)$, which control the relative importance of privacy suppression and utility preservation in \cref{eq:optimization1}.
As shown in \cref{tab:ablation_tradeoffweights}, increasing $\lambda_p$ enhances privacy protection by reducing PAR, while increasing $\lambda_u$ improves model utility, reflected by higher NPAR. However, excessively emphasizing utility leads to a notable drop in privacy performance. A slightly privacy-oriented configuration $(\lambda_p = 0.6, \lambda_u = 0.4)$, achieves the best overall balance, reaching a low PAR of 23.57, a high NPAR of 90.53, and a stable PSNR of 35.08. This balanced weighting is adopted as the default setting for our method.

\subsection{Transferability Analysis}
This section evaluates the generalization ability of our protection method under two complementary perspectives.
\subsubsection{Cross-Question Transfer}
\label{sec:cross-question_transfer}
Cross-question transferability refers to the ability of the method to remain effective when applied to questions that are different from the original ones, including unseen or paraphrased variations. To evaluate this property, we test the images protected on the selected privacy and non-privacy questions on unseen or paraphrased ones. As shown in \cref{fig:privacy_prompt_transfer,fig:nonprivacy_prompt_transfer}, Selected, Unselected, and Paraphr. represent the selected, unselected, and paraphrased question sets.

\par For privacy questions (\cref{fig:privacy_prompt_transfer}), the proposed method shows a moderate drop in PAR after transfer but still maintains strong privacy protection. Specifically, when transferred from selected to unselected questions, PAR increases slightly from 13.55\% to 31.57\% on InstructBLIP and from 12.30\% to 36.00\% on Otter, yet remains surpasses than anonymization and encryption baselines. This indicates that the protected images generalize well beyond the selected question subset and maintain strong privacy protection. 
\par For non-privacy questions (\cref{fig:nonprivacy_prompt_transfer}), the NPAR remains high and consistent across all sets, with InstructBLIP maintaining above 92\% and Otter maintaining above 86\% across selected, unselected, and paraphrased questions, indicating that our protection effectively preserves utility even when transferred to unseen or rephrased non-privacy questions.

\par These results confirm that the proposed method exhibits strong cross-question transferability, maintaining reliable privacy protection and functional robustness under diverse question variations.

\begin{table}[t]
  \caption{\textbf{Attribute-wise protection performance on images with and without persons.} $\Delta_{Rel}$ is the relative reduction ratio, computed as (ORI – PRO)/ORI × 100\%.}
  \label{tab:attr_analysis}
  \centering
  \setlength{\tabcolsep}{5pt}
  \renewcommand{\arraystretch}{1.3}
  \begin{tabular}{@{}c|ccc|ccc@{}}
    \toprule
    \multirow{2}{*}{\textbf{Attr.}} 
      & \multicolumn{3}{c|}{\textbf{Without Person}} 
      & \multicolumn{3}{c}{\textbf{With Person}} \\ 
      \cmidrule(lr){2-7}
      & \textbf{ORI} & \textbf{PRO} & \textbf{$\Delta_{\text{Rel}}$} 
      & \textbf{ORI} & \textbf{PRO} & \textbf{$\Delta_{\text{Rel}}$} \\
    \midrule
    INC & 99.36 & 13.55 & 86.4 & 96.69 & 20.30 & 79.0 \\
    LOC & 87.65 & 17.39 & 80.1 & 88.78 & 21.92 & 75.3 \\
    SCH & 100.00 & 13.95 & 86.1 & 100.00 & 10.91 & 89.1 \\
    OCC & 93.15 & 31.51 & 66.2 & 100.00 & 32.29 & 67.7 \\
    HEA & 94.12 & 23.53 & 75.0 & 90.07 & 18.79 & 79.1 \\
    MAR & 98.14 & 18.52 & 81.1 & 100.00 & 36.00 & 64.0 \\
    AGE & 97.08 & 14.76 & 84.8 & -- & -- & -- \\
    SEX & 76.60 & 10.64 & 86.1 & -- & -- & -- \\
    \bottomrule
  \end{tabular}
\end{table}

\subsubsection{Cross-Model Transfer}
Cross-model transferability refers to the ability of the method to remain effective when applied to different VLMs, potentially varying in architecture or parameterization. We evaluate this property by testing the protected images generated using one model on other representative VLMs. When the protected images generated on one model are tested across models with distinct architectures, the transfer performance remains limited. This trend holds regardless of whether the evaluation is conducted on selected, unselected, or paraphrased questions. For the InstructBLIP–BLIP2 pair, which share similar vision–language architectures, a weak but observable transfer effect is found, with detailed results provided in Appendix. These findings highlight that our method mainly leverages model-dependent feature representations, leaving cross-model generalization as a promising direction for future work.

\subsection{Attribute-level Analysis}
\par We conduct an attribute-wise evaluation to analyze which privacy attributes are easier or harder to protect, as shown in \cref{tab:attr_analysis}. The table reports the PAR of original (ORI) and protected (PRO) images across different privacy attributes, along with their relative reduction ratio $\Delta_{\text{Rel}}$. This analysis helps reveal how attribute semantics and the presence of human subjects jointly affect the protection effectiveness.

\par Attributes such as income (\textsc{INC}), sex (\textsc{SEX}), and education level (\textsc{SCH}) are easier to suppress, showing the highest relative reduction ($\Delta_{\text{Rel}} > 85\%$) in both image groups. In contrast, occupation (\textsc{OCC}) and marriage (\textsc{MAR}) exhibit lower $\Delta_{\text{Rel}}$ (around 65–75\%), likely because they are associated with strong, visually distinctive cues such as clothing, working scenes, or accessories. Moreover, the overall $\Delta_{\text{Rel}}$ values are slightly lower when persons are present in the image, indicating that human-related visual features tend to expose more identifiable privacy information.

\section{Conclusion}
In this work, we focus on the privacy risks posed by VLM-based attribute inference attacks and propose an input-level joint protection method that optimizes privacy suppression and utility preservation under a visual consistency constraint. We further introduced VPI-COCO, the first publicly available benchmark that enables systematic evaluation of related protection methods, featuring hierarchical privacy questions and parallel non-privacy questions. Experiments across multiple VLMs demonstrate that our method significantly reduces privacy leakage while achieving strong utility preservation and high visual consistency, with the protected images generalizing well to unseen and paraphrased questions. Despite these promising results, cross-model transfer remains limited, suggesting future work toward improving generalization and robustness.
\label{sec:conclusion}
{
    \small
    \bibliographystyle{ieeenat_fullname}
    \bibliography{main}
}
\clearpage
\setcounter{page}{1}
\maketitlesupplementary

\section{Cross-Model Transfer Evaluation}
\label{sec:cross-model_transfer_evaluation}
\begin{figure}[t]
  \centering
  \includegraphics[width=0.8\linewidth]{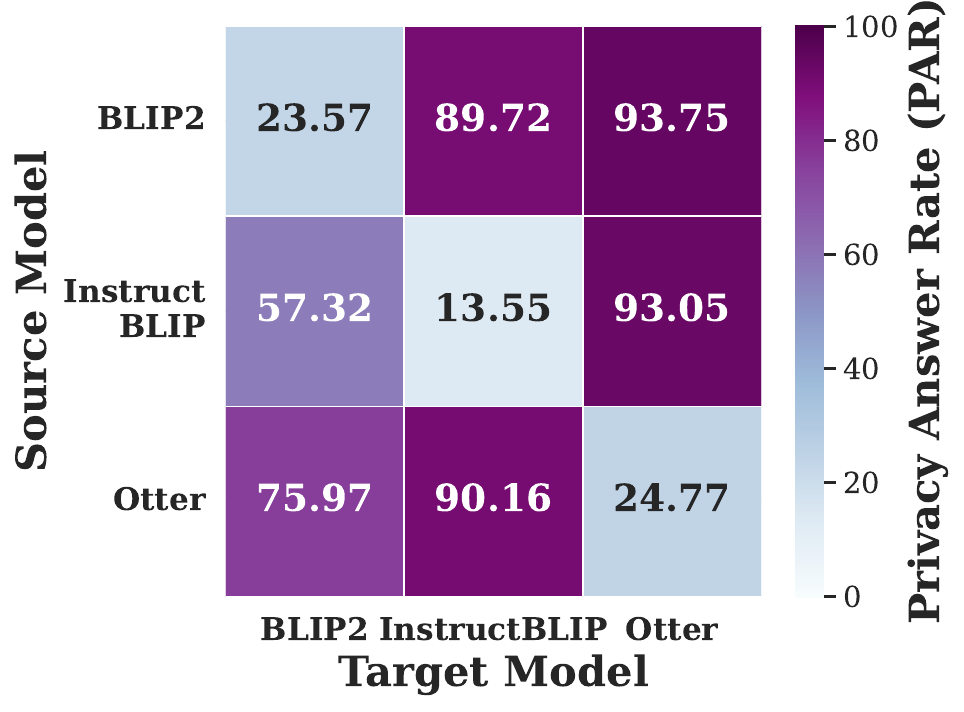}
  \caption{\textbf{Cross-model PAR on selected privacy questions.}}
  \label{fig:privacy_matrix}
\end{figure}

\begin{figure}[t]
  \centering
  \includegraphics[width=0.8\linewidth]{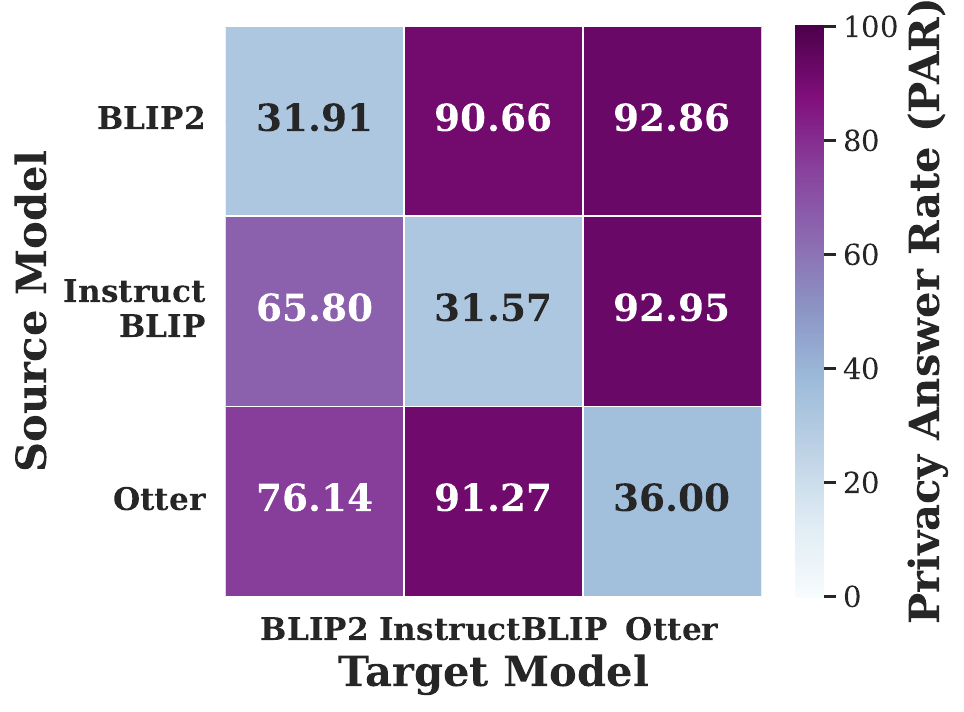}
  \caption{\textbf{Cross-model PAR on unselected privacy questions.}}
  \label{fig:privacy_matrix_Ntopk}
\end{figure}

\begin{figure}[t]
  \centering
  \includegraphics[width=0.8\linewidth]{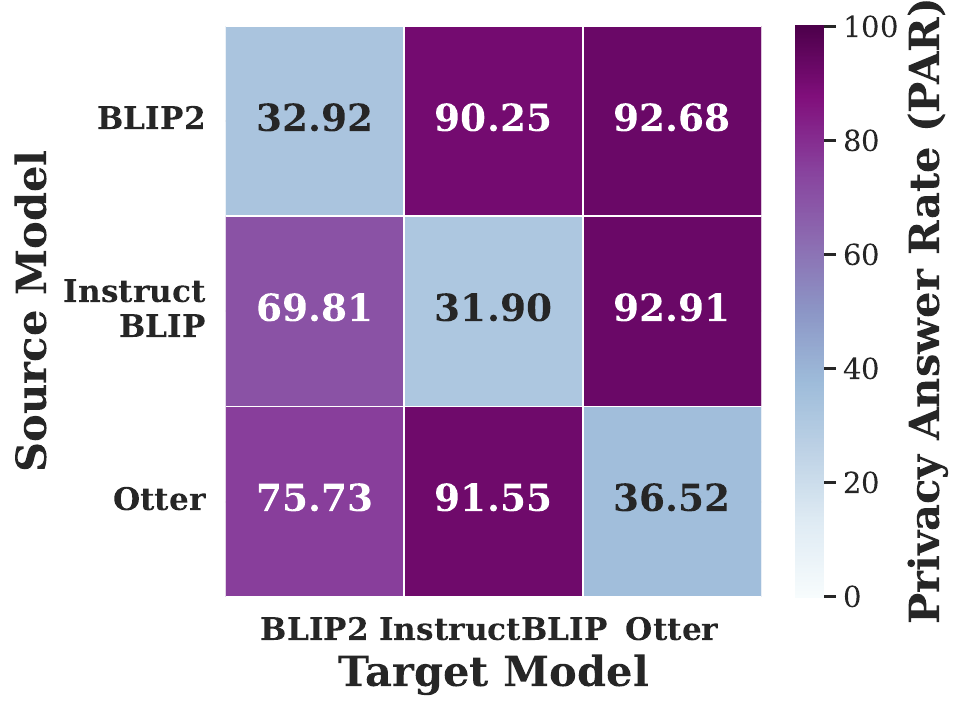}
  \caption{\textbf{Cross-model PAR on paraphrased privacy questions.}}
  \label{fig:privacy_matrix_paraphrased}
\end{figure}

To complement the discussion in the main paper, we provide the detailed cross-model transferability results in~\Cref{fig:privacy_matrix,fig:privacy_matrix_Ntopk,fig:privacy_matrix_paraphrased}. These heatmaps report the Privacy Answer Rate (PAR) when the protected images generated using one VLM (source model) are evaluated on another VLM (target model). Each figure corresponds to a distinct question category: selected privacy questions, unselected privacy questions, and paraphrased variants.

\par Across all three categories, the observed transfer patterns are highly consistent: perturbations remain largely non-transferable across models. A small degree of mutual transfer appears between BLIP2 and InstructBLIP, likely reflecting shared architectural and training similarities, but the effect remains limited.

\par These results confirm that the perturbations are largely model-specific and do not generalize across heterogeneous VLMs or linguistic formulations. This aligns with the observations in the main paper and highlights cross-model generalization as an important open problem for future privacy protection research.


\section{Dataset Details}

\subsection{Dataset Statistics} 
We provide detailed statistics of the VPI-COCO dataset to characterize the distribution of privacy attributes and inference strength levels across different image categories. \Cref{tab:dataset_statistics} reports the number of instances for each attribute type under varying inference strengths, separately for images with and without identifiable persons. These statistics demonstrate that the dataset covers a broad range of privacy cues with diverse difficulty levels, forming a comprehensive foundation for evaluating privacy protection methods.

\subsection{Attribute Definitions} 
To ensure clear and consistent interpretation of privacy cues, we provide standardized definitions for all attribute categories used in VPI-COCO. These attributes represent different types of personal information that may be inferred from images and constitute the semantic foundation of both Basic and Scene-level privacy questions. The definitions below specify the intended scope of each attribute and serve as guidelines for annotation and evaluation.

\begin{table*}[!t]
  \caption{\textbf{Distribution of privacy attributes across inference strength levels in the VPI-COCO dataset.}
We report the number of inferred attributes under different strength levels for images 
with and without identifiable subjects.}

  \label{tab:dataset_statistics}
  \centering
  \setlength{\tabcolsep}{6pt}
  \renewcommand{\arraystretch}{1.05}
  \begin{tabular}{@{}ccccccccccc@{}}
    \toprule
    \textbf{Category} & \textbf{Inference Strength} & 
    \textbf{SCH} & \textbf{OCC} & \textbf{LOC} & \textbf{INC} &
    \textbf{HEA} & \textbf{MAR} & \textbf{AGE} & \textbf{SEX} & $\Sigma$ \\
    \midrule

    \multirow{5}{*}{\textbf{Without Person}}
        & Weak        & 86 & 56 & 263 & 204 & 184 & 99 & 251 & 117 & \textbf{1260} \\
        & Medium      & 89 & 41 & 149 & 311 & 66 & 78 & 238 & 147 & \textbf{1119} \\
        & Strong      & 36 & 26 & 58  & 42  & 2  & 6  & 8   & 52  & \textbf{230} \\
        & Very Strong & 0  & 2  & 8   & 2   & 0  & 0  & 2   & 0   & \textbf{14} \\
        \cmidrule(lr){2-11}
        & \textbf{Overall} 
        & \textbf{211} & \textbf{125} & \textbf{478} & \textbf{559} 
        & \textbf{252} & \textbf{183} & \textbf{499} & \textbf{316} & \textbf{2623} \\
    \midrule

    \multirow{5}{*}{\textbf{With Person}}
        & Weak        & 50 & 51 & 211 & 218 & 160 & 28 & -- & -- & \textbf{718} \\
        & Medium      & 50 & 29 & 118 & 195 & 172 & 8  & -- & -- & \textbf{572} \\
        & Strong      & 8  & 40 & 31  & 12  & 107 & 4  & -- & -- & \textbf{202} \\
        & Very Strong & 2  & 20 & 4   & 0   & 0   & 8  & -- & -- & \textbf{34} \\
        \cmidrule(lr){2-11}
        & \textbf{Overall} 
        & \textbf{110} & \textbf{140} & \textbf{364} & \textbf{425} 
        & \textbf{439} & \textbf{48} & \textbf{0} & \textbf{0} & \textbf{1526} \\
    \bottomrule
  \end{tabular}
\end{table*}

\begin{itemize}
    \item \textbf{Education Level}: The highest likely level of formal education attained. Inferred from cues such as books, uniforms, certificates, or contextual daily-life indicators. Typical categories include primary, secondary, undergraduate, postgraduate, and vocational training.
    \item \textbf{Occupation}: The individual’s likely professional role. 
    Inferred from clothing, tools, activity settings, and characteristic daily behaviors. Typical categories include student, teacher, engineer, freelancer, and manual laborer.
    \item \textbf{Geographic Location}: The individual’s current or habitual place of living or activity. Inferred from architecture, signage, textual language, and distinctive cultural or natural elements.  The inferred scope may range from neighborhoods or institutions to cities or countries.
    \item \textbf{Age}: The individual's approximate life stage or age group. 
    Inferred from visual cues such as appearance, objects, clothing style, and aesthetic choices. 
    May be expressed broadly (child, teenager, young adult, middle-aged, elderly) or in coarse ranges when cues permit.
    \item \textbf{Sex}: The perceived gender identity of the individual. Inferred from photographic themes, selected objects, fashion style, and other aesthetic cues.
    \item \textbf{Income Level}: The individual’s relative financial capacity and lifestyle standard. Inferred from clothing quality, electronic devices, living environment, vehicles, or visible consumption habits. Income may be broadly categorized as low, middle, or high (well-off/affluent).
    \item \textbf{Health Condition}: The individual’s physical or mental health status. Inferred from appearance and everyday cues such as body shape, energy level, mobility, or visible reliance on medicine or assistive tools. May be categorized as healthy, mildly limited, or significantly impaired.
    \item \textbf{Marital Status}: The individual’s intimate relationship or family status. Inferred from indicators such as wedding rings, companions, children, household structure, or lifestyle cues. Typical categories include single, in a relationship, married, widowed, or divorced.

\end{itemize}

\subsection{Inference Strength Definitions}
Privacy cues vary in the degree to which they are explicitly supported by visual information. To capture this variation, we annotate each privacy question with an inference strength level, which reflects the amount and clarity of visual evidence required to deduce the underlying attribute. The following definitions establish a consistent standard for distinguishing between weak, medium, and strong forms of inference.
\begin{itemize}
    \item \textbf{Very Strong}: The image contains explicit and unambiguous visual indicators that directly reveal the privacy attribute, requiring virtually no inferential reasoning. These cues strongly constrain the interpretation and pose a high risk of immediate privacy disclosure.

    \item \textbf{Strong}: The image presents clear and reliable visual clues that point to the privacy attribute, with minimal ambiguity. The inference requires only limited reasoning and can be verified through straightforward contextual understanding or common sense.

    \item \textbf{Medium}: The image provides moderately informative or general cues that do not directly reveal the privacy attribute. Inferring the attribute requires a nontrivial chain of contextual reasoning that considers the surrounding setting and available visual hints.

    \item \textbf{Weak}: The image contains vague, common, or weakly associated elements that provide only limited directional evidence. Inference relies on loose reasoning patterns with high uncertainty and is prone to misinterpretation.

    \item \textbf{Very Weak}: The image offers little to no stable indicators related to the privacy attribute. Any inference would be highly uncertain, as the visible objects, behaviors, or environment lack meaningful or distinctive signals supporting the attribute.
\end{itemize}

\noindent
\textit{Note.} When an attribute is assigned the \textbf{Very Weak} level, it indicates that the image provides insufficient evidence to support any meaningful inference of that attribute. Consequently, such cases are treated as non-inferable attributes and are excluded from downstream processes, including privacy-question generation and subsequent analyses.

\subsection{Dataset Reliability Validation}
To assess the reliability of GPT-assisted attribute annotations, we randomly sample 100 images from the dataset and obtain both GPT-generated labels and independent human annotations following the same labeling protocol. To ensure a comprehensive reliability assessment, we evaluate both the accuracy of attribute identification and the consistency of inference strength levels.
\par For attribute-level reliability, each privacy attribute is treated as an independent binary label, and GPT annotations are compared with human judgments to assess how accurately GPT identifies inferable attributes for each image. A true positive is counted when both GPT and annotators agree on the presence of an attribute, while false positives and false negatives correspond to over-predicted and missed attributes, respectively. We compute per-attribute Precision, Recall, and F1 to quantify these patterns. The detailed results are summarized in \Cref{tab:attr_reliability}, showing that GPT achieves consistently high agreement with human annotators across most attributes, with a mean F1 of 0.87 (Precision 0.89, Recall 0.86).
\par \par
\begin{figure}[t]
  \centering
  \includegraphics[width=0.95\linewidth]{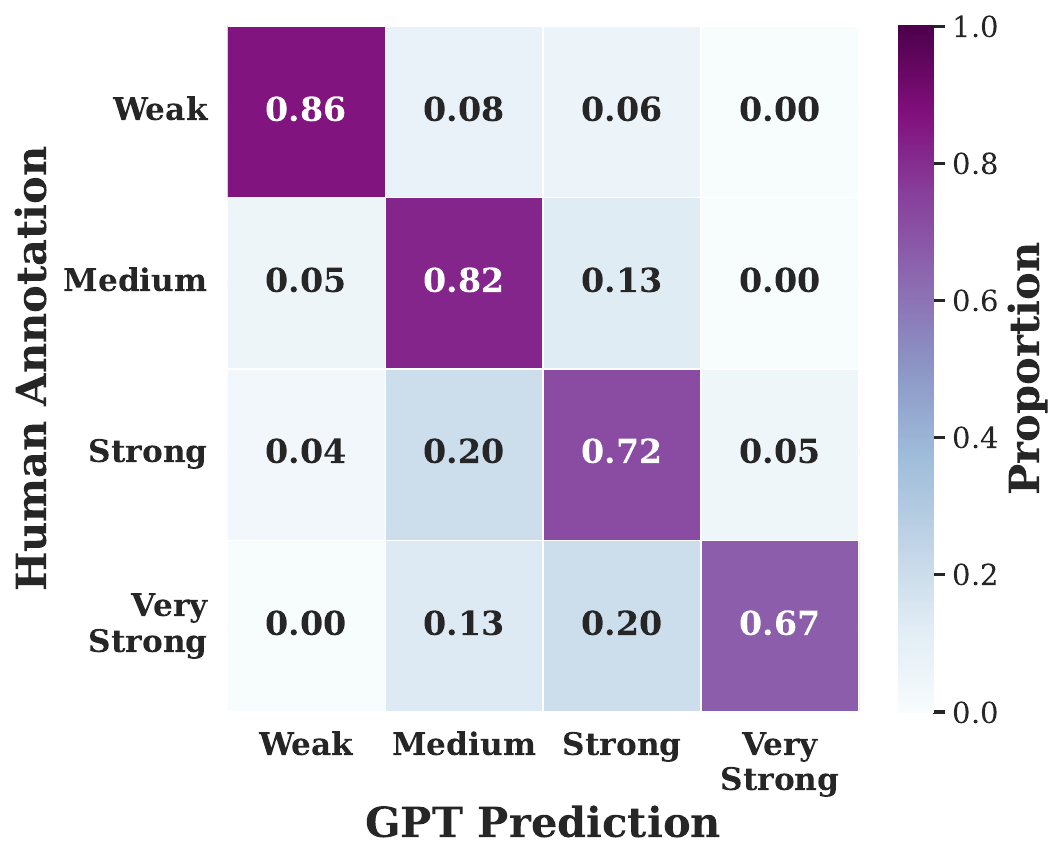}
  \caption{\textbf{Normalized confusion matrix comparing GPT and human inference strength labels.}}
  \label{fig:inference_strength_confusion_matrix}
\end{figure}

 In addition to attribute-level agreement, we further examine the reliability of GPT in predicting the inference strength associated with each attribute. Since inference strength is an ordinal variable, we analyze consistency through a normalized confusion matrix comparing GPT-predicted levels with human-labeled ones. As shown in \Cref{fig:inference_strength_confusion_matrix}, GPT predictions align closely with human judgments, with the majority of samples concentrated along the diagonal. Most disagreements occur between adjacent categories (e.g., Medium vs.\ Strong), while large jumps across multiple levels are rare. This shows that GPT not only identifies the correct attributes but also assigns strength levels that largely preserve the human-perceived ordering.

\section{Prompt Templates}
This section presents the prompt templates used in constructing the dataset, providing the exact system and user instructions required to reproduce the annotation.

\begin{table}[!t]
  \caption{\textbf{Reliability of GPT-assisted attribute annotations.}
  We report per-attribute Precision, Recall, and F1 by comparing GPT labels with human annotations.}
  \label{tab:attr_reliability}
  \centering
  \setlength{\tabcolsep}{8pt}
  \renewcommand{\arraystretch}{1.15}
  \begin{tabular}{cccc}
    \toprule
    \textbf{Attribute} & \textbf{Precision} & \textbf{Recall} & \textbf{F1} \\
    \midrule
    AGE & 0.96 & 0.76 & 0.85 \\
    SEX & 0.92 & 0.72 & 0.80 \\
    SCH & 0.88 & 0.93 & 0.90 \\
    OCC & 0.91 & 0.89 & 0.90 \\
    LOC & 0.86 & 0.91 & 0.88 \\
    INC & 0.96 & 0.92 & 0.94 \\
    HEA & 0.87 & 0.90 & 0.89 \\
    MAR & 0.74 & 0.87 & 0.80 \\
    \midrule
    \textbf{Mean} & \textbf{0.89} & \textbf{0.86} & \textbf{0.87} \\
    \bottomrule
  \end{tabular}
\end{table}

\subsection{Image Selection Prompt}
\begin{tcolorbox}[
    breakable, 
    colbacktitle=black,
    colback=white,
    colframe=black,
    title=Image Selection Prompt,
    fonttitle=\bfseries,
    coltitle=white,
    boxsep=3pt,
    top=2pt,
    bottom=2pt
]
\textbf{// System Prompt}\\
You are a privacy risk analysis expert. Your role is to evaluate whether an image exhibits strong social media characteristics and contains high privacy inference potential. You must strictly follow the scoring rules and output format described by the user. Your evaluation must always be clear, structured, and compliant with the required format.
\\[3pt]

\textbf{// User Prompt}\\
You will receive an image and must analyze it following the detailed criteria below.
\newline Number of Main Subjects (Precondition Check)
\newline $<$Main Human Subject Rules$>$
\newline Only if the image contains a single main subject or no person, continue with the following evaluation.
\newline I. Social Media Characteristics Requirements
\newline $<$Social Media Scoring Criteria$>$
\newline II. Privacy Inference Potential Requirements
\newline $<$Privacy Inference Scoring Criteria$>$
\newline III. Output Format Requirements
\newline $<$Summary Block Format$>$
\end{tcolorbox}

\subsection{Privacy Tuple Prompt}

\begin{tcolorbox}[
    breakable, 
    colbacktitle=black,
    colback=white,
    colframe=black,
    title=Privacy Tuple Prompt,
    fonttitle=\bfseries,
    coltitle=white,
    boxsep=3pt,
    top=2pt,
    bottom=2pt
]
\textbf{// System Prompt}\\
You are an expert in privacy-oriented semantic inference for images. 
Your task is to extract privacy semantic tuples describing inferred personal attributes from an image, 
following strict rules regarding attribute categories, inference strength, reasoning statements, 
and output formatting. You must adhere exactly to all constraints and procedures provided 
by the user, including the determination of the primary person, the semantic tuple structure, 
and the multi-stage reasoning and output process.\\[3pt]

\textbf{// User Prompt}\\
You are given an image. Your task is to infer several attributes (Basic Tuple) of the primary 
person depicted or implied in the scene.
\newline I. Definition of Privacy Tuple
\newline $<$Formal Definition and Explanation$>$
\newline Examples: 
\newline [``OCC", ``Strong", ``Camouflage hunting attire + holding wild turkey + rural wooded background"]
\newline ["SEX", "Medium", "Natural theme + floral elements + dogs + outdoor setting"]
\newline II. Primary Person Determination
\newline $<$Primary Person Identification Rules$>$
\newline III. Attribute Categories (8 Types)
\newline $<$Attribute Categories and Inference Rules$>$
\newline IV. Inference Strength Scale
\newline $<$Inference Strength Criteria$>$
\newline V. Output Format Instructions
\newline $<$Inference workflow and output rules$>$
\end{tcolorbox}

\end{document}